# Predicting Time-to-conversion for Dementia of Alzheimer's Type using Multi-modal Deep Survival Analysis

Ghazal Mirabnahrazam [a,1], Da Ma [a,b], Cédric Beaulac [a,c], Sieun Lee [a,d], Karteek Popuri [a,e], Hyunwoo Lee [f], Jiguo Cao [g], James E Galvin [h], Lei Wang [i], Mirza Faisal Beg [a,*], and the Alzheimer's Disease Neuroimaging Initiative [2]

[a] *School of Engineering, Simon Fraser University, Burnaby, BC, Canada*
[b] *School of Medicine, Wake Forest University, Winston-Salem, NC, USA*
[c] *Department of Mathematics and Statistics, University of Victoria, Victoria, BC, Canada*
[d] *Mental Health & Clinical Neurosciences, School of Medicine, University of Nottingham, Nottingham, United Kingdom*
[e] *Department of Computer Science, Memorial University of Newfoundland, St. John's, NL, Canada*
[f] *Division of Neurology, Department of Medicine, University of British Columbia, Vancouver, BC, Canada*
[g] *Department of Statistics and Actuarial Science, Simon Fraser University, Burnaby, BC, Canada*
[h] *Comprehensive Center for Brain Health, Department of Neurology, University of Miami Miller School of Medicine, Miami, FL, USA*
[i] *Psychiatry and Behavioral Health, Ohio State University Wexner Medical Center, Columbus, OH, USA*

## Abstract

Dementia of Alzheimer's Type (DAT) is a complex disorder influenced by numerous factors, but it is unclear how each factor contributes to disease progression. An in-depth examination of these factors may yield an accurate estimate of time-to-conversion to DAT for patients at various disease stages. We used 401 subjects with 63 features from MRI, genetic, and CDC (Cognitive tests, Demographic, and CSF) data modalities in the Alzheimer's Disease Neuroimaging Initiative (ADNI) database. We used a deep learning-based survival analysis model that extends the classic Cox regression model to predict time-to-conversion to DAT. Our findings showed that genetic features contributed the least to survival analysis, while CDC features contributed the most. Combining MRI and genetic features improved survival prediction over using either modality alone, but adding CDC to any combination of features only worked as well as using only CDC features. Consequently, our study demonstrated that using the current clinical procedure, which includes gathering cognitive test results, can outperform survival analysis results produced using costly genetic or CSF data.

* Corresponding author: Mirza Faisal Beg, PhD, P.Eng., Michael Smith Foundation for Health Research Scholar, School of Engineering Science, Simon Fraser University, ASB 8857, 8888 University Drive, Phone: (778) 782-5696, Website: http://www2.ensc.sfu.ca/ mfbeg/

Email addresses: ghazal_mirabnahrazam@sfu.ca (Ghazal Mirabnahrazam), da_ma@sfu.ca (Da Ma), cedric_beaulac@sfu.ca (Cédric Beaulac), sieun.lee@nottingham.ac.uk (Sieun Lee), kpopuri@sfu.ca (Karteek Popuri), hyunwoo.lee@ubc.ca (Hyunwoo Lee), jiguo_cao@sfu.ca (Jiguo Cao), jeg200@miami.edu (James E Galvin), Lei.Wang@osumc.edu (Lei Wang), faisal-lab@sfu.ca (Mirza Faisal Beg)

1 First Author
2 Data used in preparation of this article were obtained from the Alzheimer's disease Neuroimaging Initiative (ADNI) database (http://adni.loni.usc.edu). As such, the investigators within the ADNI contributed to the design and implementation of ADNI and/or provided data but did not participate in analysis or writing of this report. A complete listing of ADNI investigators can be found at: http://adni.loni.usc.edu/wp-content/uploads/how_to_apply/ADNI_Acknowledgement_List.pdf

# 1   Introduction

Alzheimer's Disease (AD) or Dementia of Alzheimer's Type (DAT) is a progressive neurodegenerative condition characterized by psychiatric, cognitive and structural deteriorations that accounts for 60% to 80% of all dementia cases. One out of every three seniors dies with Alzheimer's disease or another type of dementia, accounting for more deaths than breast and prostate cancer combined (Alzheimer's Association 2021). Since there is no cure available for AD, there is a substantial interest in finding ways to better understand the characteristics of the disease and to develop methods that can successfully detect those at risk at an early stage of the disease before symptomatic onset.

Many factors contribute to the development and progression of Alzheimer's disease, but the extent to which each factor affects the disease is still unknown. As a result, it is critical to thoroughly investigate the effects of phenotype, genotype, and lifestyle factors on the development and progression of DAT. Data from various modalities were obtained and analyzed in the search for biomarkers that can accurately diagnose DAT in its early stages. Magnetic resonance imaging (MRI), for example, is the most widely used data modality for identifying specific structural changes in terms of atrophy in the brain associated with DAT progression (Hua et al. 2008; Vemuri and Jack 2010; Popuri et al. 2020). Cerebrospinal Fluid (CSF) biomarkers such as abnormal amyloid and tau levels can reflect the intensity of disease progression and have been extensively studied in relation to Alzheimer's disease (Olsson et al. 2016; Finehout et al. 2007; Anoop et al. 2010). Another modality that has been shown to be effective in predicting the likelihood of developing DAT even before pathological changes begin is genetic information. A number of genetic risk factors have been linked to DAT, with the APOE-$\varepsilon$4 allele accounting for 20-25 percent of cases (Lambert et al. 2013). Multiple genome-wide association studies (GWAS) have also found possible links between Single Nucleotide Polymorphisms (SNPs) and DAT (Jansen et al. 2019; Kunkle et al. 2019; Schwartzentruber et al. 2021). At the time of writing this manuscript, 20 genes had been reported to be associated with AD via GWAS, the majority of which were associated with moderate to small effect sizes (Lambert et al. 2013). Other factors studied alone or in combination with other modalities to predict Alzheimer's disease progression include socio-demographic and clinical data, as well as cognitive performance tests (Grassi et al. 2019; Lei et al. 2020; Devanand et al. 2008). The growing availability of databases containing multiple data modalities, such as the Alzheimer's Disease Neuroimaging Initiative (ADNI), has allowed researchers to investigate the effects of integrating multi-modal data in DAT risk prediction (Venugopalan et al. 2021; An et al. 2017; Zhou et al. 2019). While it has been demonstrated that combining different data modalities improves diagnosis performance, there is still a lack of understanding about how each modality contributes to DAT diagnosis, and translation into practice is still limited.

Early diagnosis is critical for successful disease management and the ability to use disease-modifying drugs to alleviate symptoms. In current clinical practice, a DAT diagnosis cannot be made until the patient exhibits clear signs of cognitive decline, which can be attributed in part to the multifactorial nature of DAT. Methods for predicting the probability of a patient developing Alzheimer's disease as a function of time, known as survival analysis, are important tools in helping understand the characteristics of DAT. One of the most important contributions of survival analysis methods is that they can account for individuals who are not followed up to their dementia onset time, i.e., censored individuals, allowing them to utilize and provide more information than the traditional classification methods. In addition, it is necessary to forecast whether a subject would convert to dementia, and when, specifically, the conversion would occur. While classification methods can solve the first problem, they are not suitable for predicting the time to conversion to DAT, particularly in the presence of censored individuals. Therefore, survival analysis is a more appropriate method, when the time to the diagnosis of dementia may be unknown.

The application of survival analysis methods to predict the time to conversion of Alzheimer's disease is relatively limited, and the available studies in the literature have mostly focused on using single data modalities as predictive features (Nakagawa et al. 2020; Orozco-Sanchez et al. 2019; Aschwanden et al. 2020), or only on predicting the time to DAT conversion for patients with Mild Cognitive Impairment

(MCI), or time to conversion to MCI for healthy subjects rather than predicting the conversion time for subjects in all stages of the disease (Pölsterl et al. 2019; Spooner et al. 2020; Lu and Colliot 2021). To the best of our knowledge, this is the first study that performs a comprehensive analysis on the prediction of the time to conversion to DAT for subjects in various stages of the disease using multi-modal data and compares the predictive power of each data modality and the effect of each modality on the disease diagnosis and progression.

## 2  Material

### 2.1  Data

Data used in preparation of this study was obtained from the publicly available Alzheimer's Disease Neuroimaging Initiative (ADNI) database (http://adni.loni.usc.edu). The ADNI was launched in 2003 as a public-private partnership, led by Principal Investigator Michael W. Weiner, MD. The primary goal of ADNI has been to test whether serial MRI, PET, other biological markers, and clinical and neuropsychological assessment can be combined to measure the progression of mild cognitive impairment (MCI) and early Alzheimer's disease (AD). In addition, ADNI aims to provide researchers with the opportunity to combine genetics with imaging and clinical data to help investigate mechanisms of the disease.

### 2.2  Data selection and stratification

A total of 401 subjects from the first phase of ADNI (ADNI1; Mueller et al. 2005) who had the following 5 data modalities available at baseline were included in the study:

1. MRI data;
2. Genetic data including Single Nucleotide Polymorphism (SNP) + APOE information (GEN);
3. Cognitive tests (COG) such as the Mini Mental State Examination (MMSE);
4. Demographic data (DEM) such as age, sex, education and marital status;
5. Cerebrospinal fluid (CSF) data.

A data stratification method with a focus on the subjects' past, current and future clinical diagnosis (Popuri et al. 2018; 2020; Mirabnahrazam et al. 2022) was used to divide the subjects into five subgroups. Based on the information available during the ADNI study period, each participant was assigned to one of the five subgroups described in Table 1. Additional detail about the data stratification method can be found in our previous publication (Mirabnahrazam et al. 2022). Subjects from the stable(s)NC, unstable(u)NC, progressive(p)NC, stable(s)MCI, and progressive(p)MCI stratified subgroups were included in this study. These stratified subgroups were divided into two categories based on whether they received a clinical diagnosis of DAT in a future time point or not. The group descriptions are as follows:

1. Non-progressive (right-censored): This group includes subjects from the sNC, uNC, and sMCI stratified groups who did not receive a clinical diagnosis of DAT throughout the study period.
2. Progressive (uncensored): This group includes subjects from the pNC and pMCI groups who received a DAT diagnosis after their initial visit during the study window.

The predict labels for each subject in the study include: a) an event indicator of 0 or 1 indicating whether the subject was progressive (1) or non-progressive (0), and b) a duration indicating the time between the first visit and the time when the DAT diagnosis was confirmed for progressive subjects, and the time between the first and last visit for non-progressive subjects.

## 2.3 Feature pre-processing for the Multi-modal input data

We used all of the available features from the Cognitive tests (10 features), Demographic (4 features), and CSF (7 features) categories in the ADNI database, and created a new set of 21 features called CDC. Then, in order to perform a fair comparison between feature sets, we selected the top 21 most important features from MRI and genetic data modalities. More details of the feature selection process were described in detail in our previous work (Mirabnahrazam et al. 2022). All 63 features chosen for this study are listed in Table A.1, available in Section A of the Supplementary Material.

**Table 1.** Demographic information and progression group division for subjects included in the study. Each subject is assigned a membership in the form of `prefix{Group}', where `Group' is the clinical diagnosis at baseline, and `prefix' signals future clinical diagnoses. The stratified subgroups were divided into two groups based on whether or not they received a clinical diagnosis of DAT in a future time point. The non-progressive group includes subjects from the sNC, uNC, and sMCI stratified groups who did not receive a clinical diagnosis of DAT during the study period, and the progressive group includes subjects from the pNC and pMCI groups who received a DAT diagnosis after their initial visit during the study window.

| Dementia trajectory | Group name | Clinical diagnosis at baseline | Clinical progression | Subjects [M:F] | Age [c] [Years] | CSF [a,c] [t-tau/Aβ1-42] |
|---|---|---|---|---|---|---|
| Non-progressive [b] | sNC: stable NC | NC [a] | **NC** [d] → **NC** | 58:51 | 75.79 (4.93) | 0.34 (0.23) |
| Non-progressive | uNC: unstable NC | NC | **NC** → MCI | 14:8 | 76.57 (3.70) | 0.39 (0.19) |
| Non-progressive | sMCI: stable MCI | MCI [a] | **MCI** [d] → **MCI** | 65:36 | 74.70 (7.35) | 0.67 (0.52) |
| Progressive [b] | pNC: progressive NC | NC | **NC** → MCI → DAT | 6:8 | 76.49 (4.33) | 0.75 (0.42) |
| Progressive | pMCI: progressive MCI | MCI | **MCI** → DAT | 99:56 | 73.85 (6.85) | 0.82 (0.45) |

[a] NC: normal control, MCI: mild cognitive impairment, DAT: dementia of Alzheimer's type, CSF: cerebrospinal fluid, t-tau: total tau, Aβ1-42: beta amyloid 1-42,

[b] Non-progressive: right-censored, subjects who did not receive a DAT diagnosis within the study window, Progressive: uncensored, subjects who received a diagnosis of DAT during the study window,

[c] The mean (standard deviation) age and CSF measure values within each group are given; CSF measures were only available for a subset of subjects in each of the groups: sNC (57), uNC (17), sMCI (55), pNC (8), pMCI (88),

[d] Clinical diagnosis at baseline is shown in **bold** under the "Clinical progression" column.

To avoid removing subjects who were missing some information from the 63 features chosen above, the missing values for each feature were replaced by an out-of-range value as suggested by Twala & al. (Twala, Jones, and Hand 2008). Section B in Supplementary Material includes a comparison between this method and two other methods for handling missing values. Feature scaling has been performed on the data as a pre-processing step to ensure a consistent set of feature ranges (Ma et al. 2019). Standardization ($\hat{x} = \frac{x-\bar{x}}{\sigma}$) was performed on features with either categorical or continuous data types, while binary features remained unchanged. $\hat{x}$ is the standardized feature vector, $x$ is the original feature vector, $\bar{x}$ is the mean of that feature vector, and $\sigma$ is its standard deviation. Section C in Supplementary Material compares the performance of our pre-processing method with two other methods.

## 3 Methods

### 3.1 DeepSurv survival model

Survival analysis, or time-to-event analysis, is used to estimate the time until an individual or a group of individuals experience an event of interest. It is common in time-to-event data that some individuals are not followed up to their event time for a variety of reasons, including leaving the study or experiencing another event that prevents them from experiencing the event of interest, resulting in censored times rather than event times. While many analyses ignore these observations, one of the most important contributions of survival analysis methods is to account for them.

Survival analysis has received substantial recent attention in the machine learning literature. The application of neural networks to survival analysis was first introduced by (Faraggi and Simon 1995), where they extended the classical Cox proportional hazard model (Cox 1972) by using a one hidden layer multilayer perceptron (MLP) to learn the relationship of the covariates to the hazard function. Recent advances in deep learning have enabled researchers to develop several cutting-edge deep learning based survival analysis models that can overcome the constraints of the traditional models such as the linearity assumption between the covariates and the hazard function. The models have been shown to be more successful at accurately estimating the underlying relationship between covariates and the event of interest in complex problems than the traditional models.

We have utilized a deep learning based survival analysis model called DeepSurv (Katzman et al. 2018) which extends the classic Cox proportional hazard model to predict, analyze, and compare the time-to-conversion to DAT. Cox regression model (Cox 1972) provides a semi-parametric specification of the hazard rate:

$$h(t|x) = h_0(t)exp[g(x)], \quad g(x) = \beta^T x, \quad (1)$$

where $h_0(t)$ is a non-parametric baseline hazard, and $\exp[g(x)]$ is the hazard ratio or risk score. Here, $x$ is a covariate vector or a vector of features included in the study and $\beta$ is a parameter vector. The hazard ratio is the parametric part of the model which consists of a linear predictor $g(x) = \beta^T x$. The Cox partial likelihood, with Breslow's method for handling tied event times (Breslow 1972), is given by:

$$L_{cox} = \prod_i \left( \frac{exp[g(x_i)]}{\sum_{j \epsilon R_i} exp[g(x_j)]} \right)^{D_i}, \quad (2)$$

and the negative partial log-likelihood can then be used as a loss function:

$$loss = \sum_i D_i \log \left( \sum_{j \epsilon R_i} exp[g(x_j) - g(x_i)] \right), \quad (3)$$

where $i$ denotes an individual, $D_i$ is an indicator labelling the observed event time as an progressive (uncensored; $D_i = 1$) or non-progressive (right-censored; $D_i = 0$) observation, and $R_i$ is the set of all individuals at risk at time $T_i$ (not censored and have not experienced the event before time $T_i$). The negative partial log-likelihood is usually minimized using Newton-Raphson's method.

To construct a non-linear version of the Cox model, the linear predictor $g(x) = \beta^T x$ in the relative risk function above is replaced by a $g(x)$ parametrized by a neural network. The predictions are obtained by estimating the survival function, $\hat{S}(t|x) = \exp[-\hat{H}(t|x)]$, where $\hat{H}(t|x)$ is the cumulative hazard function, which is commonly used for specifying different survival models, and is defined as:

$$H(t|x) = \int_0^t h(s|x)\,ds = \int_0^t h_0(s)\,exp[g(x)]\,ds, \qquad (4)$$

and in practice can be estimated by:

$$\widehat{H}(t|x) = \sum_{T_i \le t} \Delta \widehat{H_0}(T_i)\,exp[\hat{g}(x)], \qquad (5)$$

where $\Delta \widehat{H_0}(T_i)$ is an increment of the Breslow estimate (Breslow 1972):

$$\Delta \widehat{H_0}(T_i) = \frac{D_i}{\sum_{j \epsilon R_i} exp[\hat{g}(x_j)]}, \qquad (6)$$

and $\hat{g}(x)$ is the estimate of $g(x)$ obtained from the neural network.

In order to evaluate the performance of the above model, we have trained our data using the Cox model (Cox 1972) as benchmark, as well as four other best-performing deep learning-based models in the literature and reported the results in section D of Supplementary Material.

## 3.2 Evaluation metrics

### 3.2.1 Concordance index

Concordance index or C-index (Harrell et al. 1982) is arguably the most widely used metric for global assessment of prognostic models in survival analysis. The C-index estimates the likelihood that the predicted survival times of a random pair of individuals will have the same ordering as their true survival times among all pairs of subjects that can be ordered. The C-index attempts to describe the performance of a model based on the assumption that patients who lived longer should be assigned a lower risk than patients who lived shorter. A "good" model, according to the C-index (C=1), is the one that always assigns higher scores to the subjects who have experienced the earlier events.

Figure 1 illustrates the graphical representation of C-index computation in the presence of censored data inspired by a graphical representation proposed by Steck et al. (2008). When calculating the C-index between two data points, we can determine the order of events if both data points are progressive (uncensored). If one of the data points is non-progressive (right-censored), concordance can be calculated only if the censoring occurs after the event for the progressive (uncensored) data point. Concordance cannot be evaluated for a pair if both data points are non-progressive (right-censored) or if both events occur at the same time.

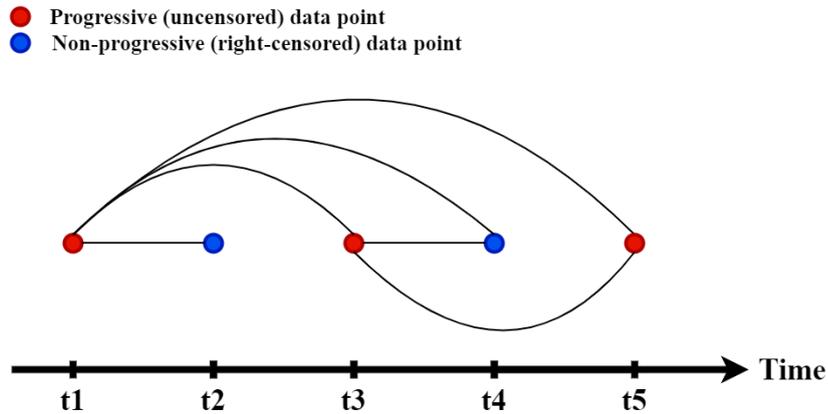

**Figure 1.** **Graphical representation of C-index computation. Each circle indicates the predicted survival time for a data point. The red circles represent the progressive (uncensored) data points and the blue circles represent the non-progressive (right-censored) data points. The figure illustrates the pairs of data points for which an order of events can be established.**

The C-index is computed at the initial time of observation and only depends on the ordering of the predictions, hence it cannot reflect the possible change in the risk over time. Therefore, here we use the time-dependent concordance index (Ctd-index) (Antolini, Boracchi, and Biganzoli 2005), which estimates the probability that observations $i$ and $j$ are concordant provided that they are comparable:

$$C^{td} = P\{\hat{S}(T_i \mid x_i) < \hat{S}(T_j \mid x_j) \mid T_i < T_j,\ D_i = 1\}. \qquad (7)$$

### 3.2.2 Brier score

The Brier score (BS) (Brier 1950) is used to evaluate the accuracy of a predicted survival function; it represents the average squared distances between the observed survival status, $y_i \in \{0,1\}$, and the predicted survival probability, $\hat{p}_i$ and is always a number between 0 and 1, with 0 being the best possible value.

$$BS = \frac{1}{N} \Sigma_i (y_i - \hat{p}_i)^2, \qquad (8)$$

where $N$ is the number of observations. To get binary outcomes from time-to-event data, we choose a fixed time $t$ and label data according to whether or not an individual's event time is shorter or longer than $t$. To account for censored data, the Brier score have been generalized (Graf et al. 1999) by re-weighting the scores by the inverse censoring distribution,

$$BS(t) = \frac{1}{N} \sum_{i=1}^{N} \left[ \frac{\hat{S}(t|x_i)^2 \mathbb{1}\{T_i \leq t, D_i = 1\}}{\hat{G}(T_i)} + \frac{(1-\hat{S}(t|x_i))^2 \mathbb{1}\{T_i > t\}}{\hat{G}(t)} \right]. \quad (9)$$

Here N is the number of observations, $\hat{G}(T_i)$ and $\hat{G}(t)$ are the Kaplan-Meier (Kaplan and Meier 1958) estimates of censoring distribution at times $T_i$ and t, and it is assumed that the censoring times and survival times are independent. The Brier score can be extended from a single duration $t$ to an interval by computing the integrated Brier score (IBS):

$$IBS = \frac{1}{t_2 - t_1} \int_{t_1}^{t_2} BS(s) \, ds. \quad (10)$$

## 3.3 Network Architecture

The base neural network used to train the data is a multilayer perceptron (MLP) with the same number of nodes in each hidden layer, rectified linear unit (ReLU) activation function, and batch normalization between layers. For regularization, we used dropout (Srivastava et al. 2014), normalized decoupled weight decay (Loshchilov and Hutter 2017) and early stopping. We utilized the cyclic AdamWR (Loshchilov and Hutter 2017) optimizer with an initial cycle length of one epoch. The optimizer multiplies the learning rate with 0.8 and doubles cycle length after every cycle. The initial learning rate was found using the methods proposed by Smith (2017). The output of the network is a single node that estimates the hazard rate in the Cox model (equation (1)). We then obtain the predictions by estimating the survival function, $\hat{S}(t|x) = \exp[-\hat{H}(t|x)]$, from the estimated cumulative hazard by following equations (4 - 6).

We performed hyperparameter search using the repeated random sub-sampling approach, also known as the Monte Carlo cross-validation (Xu and Liang 2001), with 100 splits on our data including all 63 features. In each split, 80% of the subjects were chosen at random for training, with the remaining 20% reserved for validation. The data was stratified so that each time, 80% of the subjects from the sNC, uNC, pNC, sMCI, and pMCI groups were included in the training set. This method ensures that the network is exposed to a representative subset of data in each split, resulting in a more accurate learning experience. We ran hyperparameter search on the parameters listed in Table 2 and chose the model with the highest validation set score calculated from the loss function in equation (3). The hyperparameter values highlighted in **bold** in Table 2 were used as a fixed set of parameters to train the data using different feature combinations across all experiments. Figure 2 illustrates our network architecture.

**Table 2.** Hyperparameter search space for model optimization.

| Hyperparameter | Values |
| --- | --- |
| Hidden layers | {1, 2, **3**[a], 4, 5, 6} |
| Nodes per layer | {10, 25, 32, 50, 64, **75**, 100} |
| Dropout | {0, 0.1, 0.2, **0.3**, 0.4, 0.5, 0.6, 0.7} |
| Weight decay | {0, **0.01**, 0.02, 0.05, 0.1, 0.2, 0.5} |
| Batch size | {**16**, 32, 64, 128, 256, 512} |

[a] Hyperparameter values used as the fixed parameters in the network architecture are highlighted in **bold**.

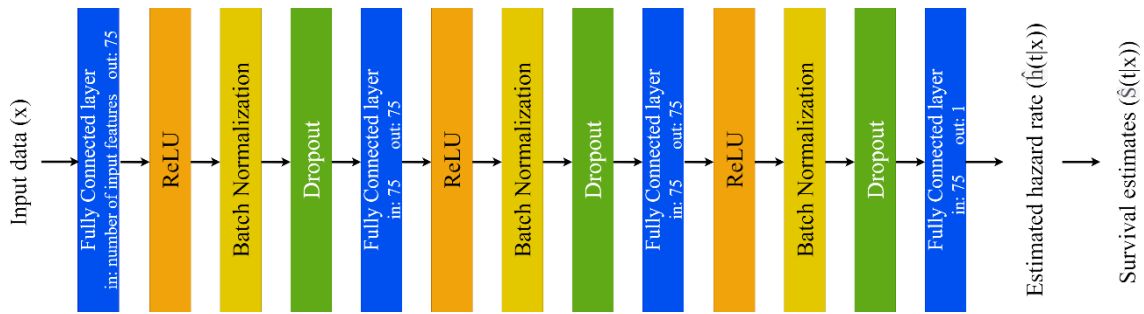

**Figure 2.** **Network architecture used as the base model. Input data includes a subject's feature vector (x), including MRI, genetic, or CDC features and the network's output is the estimated hazard rate for that subject. The survival estimate for a subject is calculated from the estimated hazard rate using $\widehat{S}(t|x) = \exp[-\widehat{H}(t|x)]$ where $\widehat{H}(t|x)$ is the cumulative hazard rate.**

## 3.4 Experiments

### 3.4.1 Multi-modal survival comparison

Using Monte Carlo cross-validation (Xu and Liang 2001) with 10 splits, each time we randomly selected 80% of the subjects for training and 20% of the subjects for testing. The train-test split has been performed in a stratified fashion such that each time 80% of the subjects in each of the sNC, uNC, pNC, sMCI, and pMCI were included in the training set. In each split, 20% of the training subjects were randomly selected for internal validation. The number of subjects included in each of the 10 training, validation, and testing sets is shown in Table 3.

**Table 3.    Number of subjects included in the training, testing, and validation sets.**

| Groups | Training set | Validation set | Testing set | Total |
|---|---|---|---|---|
| sNC | 70 | 17 | 22 | 109 |
| uNC | 14 | 4 | 4 | 22 |
| pNC | 9 | 2 | 3 | 14 |
| sMCI | 65 | 16 | 20 | 101 |
| pMCI | 99 | 25 | 31 | 155 |
| Total | 257 | 64 | 80 | 401 |

Seven experiments were conducted using identical hyperparameters for the model using the following features:

1. Genetic data only (GEN; 21 features),
2. MRI data only (MRI; 21 features),
3. The combination of cognitive test data, demographic, and CSF measures (CDC; 21 features),
4. Combined MRI and genetic data (GEN+MRI; 42 features),
5. Combined genetic and CDC data (GEN+CDC; 42 features),
6. Combine MRI and CDC data (MRI+CDC; 42 features),
7. All features combined (GEN+MRI+ CDC; 63 features).

In each split, our method estimates the survival rate for each subject over a 10-year period. To evaluate and compare the performance of the model, we reported the average of time-dependent concordance index ($C^{td}$-index) and the Integrated Brier score (IBS) over 10 splits.

### 3.4.2 Feature importance analysis

To assess the significance of the features used in the previous step, we used a method called Permutation Importance (Breiman 2001), which allows us to treat and inspect fitted machine learning models as black-box estimators. The idea behind the technique is to compute the feature importance by measuring how the evaluation score decreases when a feature is not available. This is accomplished by shuffling the values in the original feature to generate random noise with the same distribution as the original feature. Because this procedure breaks the link between the feature and the target, the drop in the model score reflects how much the model is dependent on the feature.

The importance of each feature $j$ (a column of the dataset) is defined as:

$$i_j = s - \frac{1}{K} \sum_{k=1}^{K} s_{k,j} \quad (11)$$

where $s$ is the reference score of the trained model on the unshuffled data, $K$ is the number of times column $j$ was shuffled, and $s_{k,j}$ is the calculated score on the data including the shuffled column $j$ at iteration $k$. Column $j$ is randomly shuffled $K$ times, and the resulting scores are averaged; the difference between the reference score and the averaged scores from the shuffled data is then defined as the importance of feature $j$. Positive feature importance ($i_j > 0$) denotes a decrease in the score in the absence of feature $j$, and thus the importance of feature $j$, while negative feature importance ($i_j < 0$) denotes an increase in the score in the absence of the feature and the potential negative effect the feature can have on the performance of the model.

We computed feature importance on the GEN, MRI, CDC, and GEN+MRI+CDC feature sets to investigate the significance of each feature type alone and in combination. $C^{td}$-index have been used as the evaluating score, and each of the features were shuffled 10 times. The scores were calculated on the trained models from step 1 and using the data in the testing sets.

Using the results of feature importance above, we removed features with negative feature importance ($i_j < 0$) from each of the MRI, GEN, CDC, and MRI+GEN+CDC feature sets and retained the models following the same settings as before to investigate whether including those features reduces the overall performance of the model.

### 3.4.3 Progressive vs. non-progressive survival analysis

To compute the survival estimates over time, in each split, we used our trained model to estimate the survival curve for 80 subjects in the testing set. Next, we averaged the survival curves of subjects who were in more than one testing set to create the final set of survival estimates across 10 splits. Some of the subjects were never included in the testing set. Table 6 shows the number of subjects who were included in the testing set at least once across 10 splits.

A good survival model should predict a high survival rate for non-progressive subjects, whereas the progressive subjects should be associated with a low survival rate. To validate the above statement, we used the survival estimates of the 206 non-progressive and 150 progressive subjects obtained using all available features (MRI+GEN+CDC), and examined the survival trend of subjects in these two groups. We investigated the survival trend over 1, 2, 5, and 10 year durations, starting from the initial clinical visit for each subject.

**Table 4.** **Number of subjects included in the testing set at least one time in 10 splits. Our trained model was used to estimate a survival curve for these subjects.**

| Stratified group | sNC | uNC | sMCI | pNC | pMCI |
|---|---|---|---|---|---|
| # subjects (testing set/total) | 96/109 | 20/22 | 90/101 | 13/14 | 137/155 |
| Group | Non-progressive | | | progressive | |
| # sum (testing set/total) | 206/232 | | | 150/169 | |

# 4  Results

## 4.1  Multi-modal survival comparison

Figure 3 illustrates performance comparison of the seven different feature sets used to train our model. The top figure represents the $C^{td}$-index performance results, where higher scores indicate superior performance, and the bottom figure shows the IBS results, where lower scores are preferable. When comparing the results obtained from a single modality (GEN, MRI, and CDC), CDC (green) showed to produce a significantly better performance using both $C^{td}$-index (0.822) and IBS (0.111) compared to GEN ($C^{td}$-index p-value: $2 \times 10^{-7}$; IBS p-value: $1.38 \times 10^{-5}$) and MRI ($C^{td}$-index p-value: $2.18 \times 10^{-6}$; IBS p-value: $2.45 \times 10^{-5}$). Combining MRI and GEN features (red) improved model performance over using either GEN (blue) or MRI (orange) features alone using both metrics ($C^{td}$-index: 0.76 vs. 0.589 and 0.727; IBS: 0.148 vs. 0.206 and 0.16). This improvement was statistically significant using both metrics when compared to GEN ($C^{td}$-index p-value: $3.57 \times 10^{-5}$; IBS p-value: 0.0017), and using $C^{td}$-index when compared to MRI (p-value: 0.0334).

Adding CDC to any feature set (GEN+CDC (purple), MRI+CDC (brown), and MRI+GEN+CDC (pink)) improved performance over using that feature set without CDC (GEN (blue), MRI (orange), MRI+GEN (red)). However, when compared to using CDC features alone, combining CDC with other features did not always result in an improved performance. The combination of MRI and CDC features (brown) had the best overall performance, closely followed by the results obtained using only CDC features ($C^{td}$-index: 0.831 (MRI+CDC) and 0.822 (CDC); IBS: 0.105 (MRI+CDC) and 0.111 (CDC)).

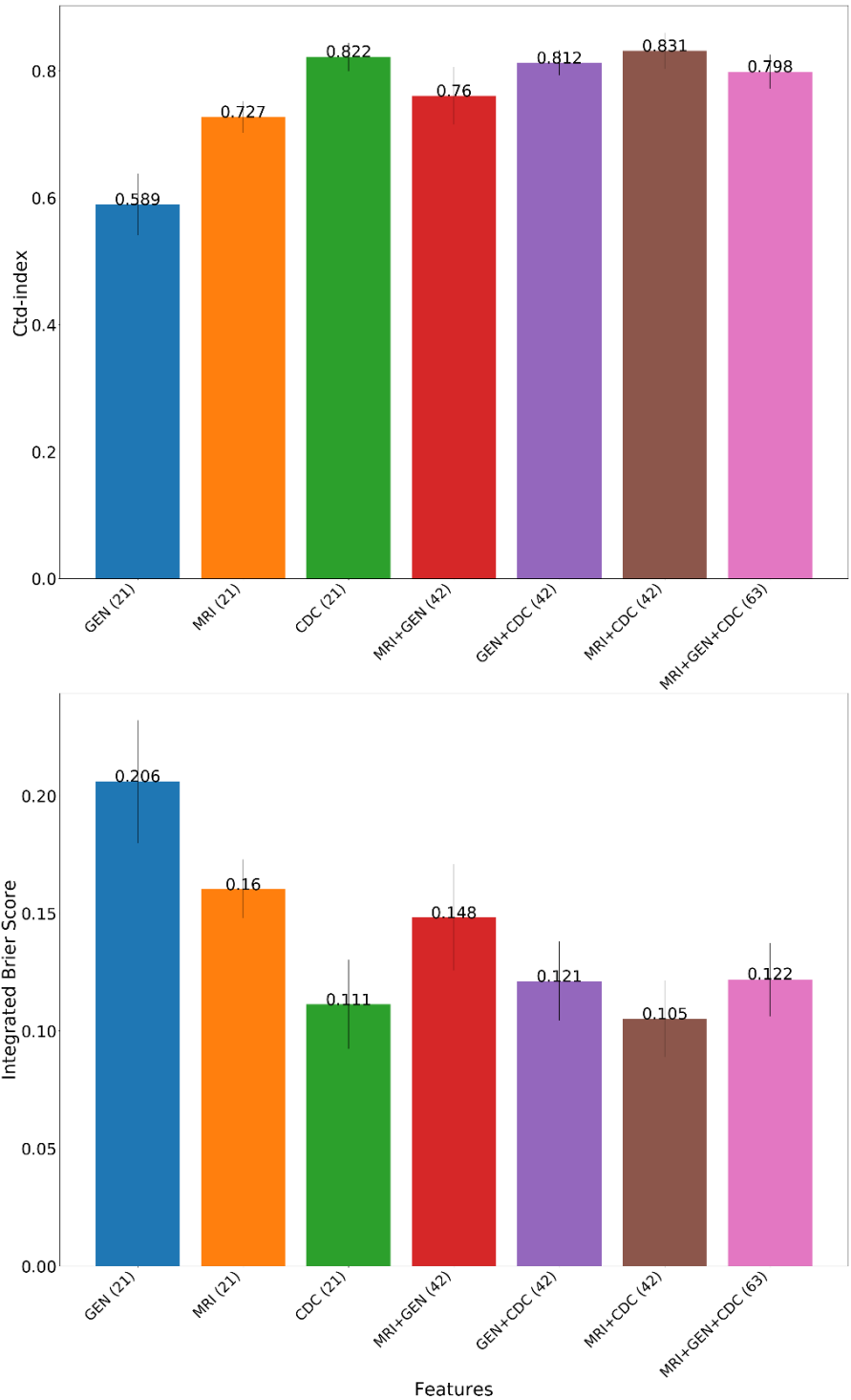

**Figure 3.** **Performance comparison between different feature sets using time the dependant concordance index (Top row), and Integrated Brier score (Bottom row). Each bar represents the mean C-index over ten splits, with the standard deviation indicated by a vertical error bar. The mean value has been printed above each bar. The name of each feature set (number of features) has been shown on x asis.**

## 4.2  Feature Importance analysis

Figure 4 shows the results of feature importance analysis for models train using GEN, MRI and CDC feature sets. The top graph depicts the feature importance rates for GEN features. 14 of the 21 features in the features set have been shown to have a positive effect on performance. The most important feature in this feature set is the well-known AD biomarker APOEε4, which is followed by rs2883782 on chromosome 2 and rs10510985, rs7627954, and rs6773506 on chromosome 3. The middle graph in Figure 4 shows feature importance results for models trained with MRI features. 15 of the 21 features used were shown to have a positive effect on the performance. The Hippocampus region on the left hemisphere is the most important feature in this feature set and its importance rate is at least twice bigger than the importance rate for all other features. Other important features include Amygdala (left hemisphere), Hippocampus (right hemisphere), Inferior parietal lobule (right hemisphere), and Supramarginal gyrus (left hemisphere). The feature importance results for models trained with CDC features are shown in the bottom graph of Figure 4. The majority of the features in this features set were found to be important, with only 2 showing negative importance rates. 8 of the top 10 most important features were drawn from cognitive tests, with the Delayed recall variable from the Logical memory test (LDEL-Total) being the most important. Sex and education were also among the top 10.

The result of feature importance analysis on models trained with all available features (GEN+MRI+CDC) is shown in Figure 5. We found that 27 of the 63 features, including 6 GEN features, 9 MRI features, and 12 CDC features, were shown to have a positive effect. The top five most important features were from cognitive test features in the CDC feature modality, with the CDRSB (Clinical Dementia Rating Scale) being the most important, closely followed by LDEL-Total. The Hippocampus on the left hemisphere is the most important MRI feature, ranking among the top ten in this feature set.

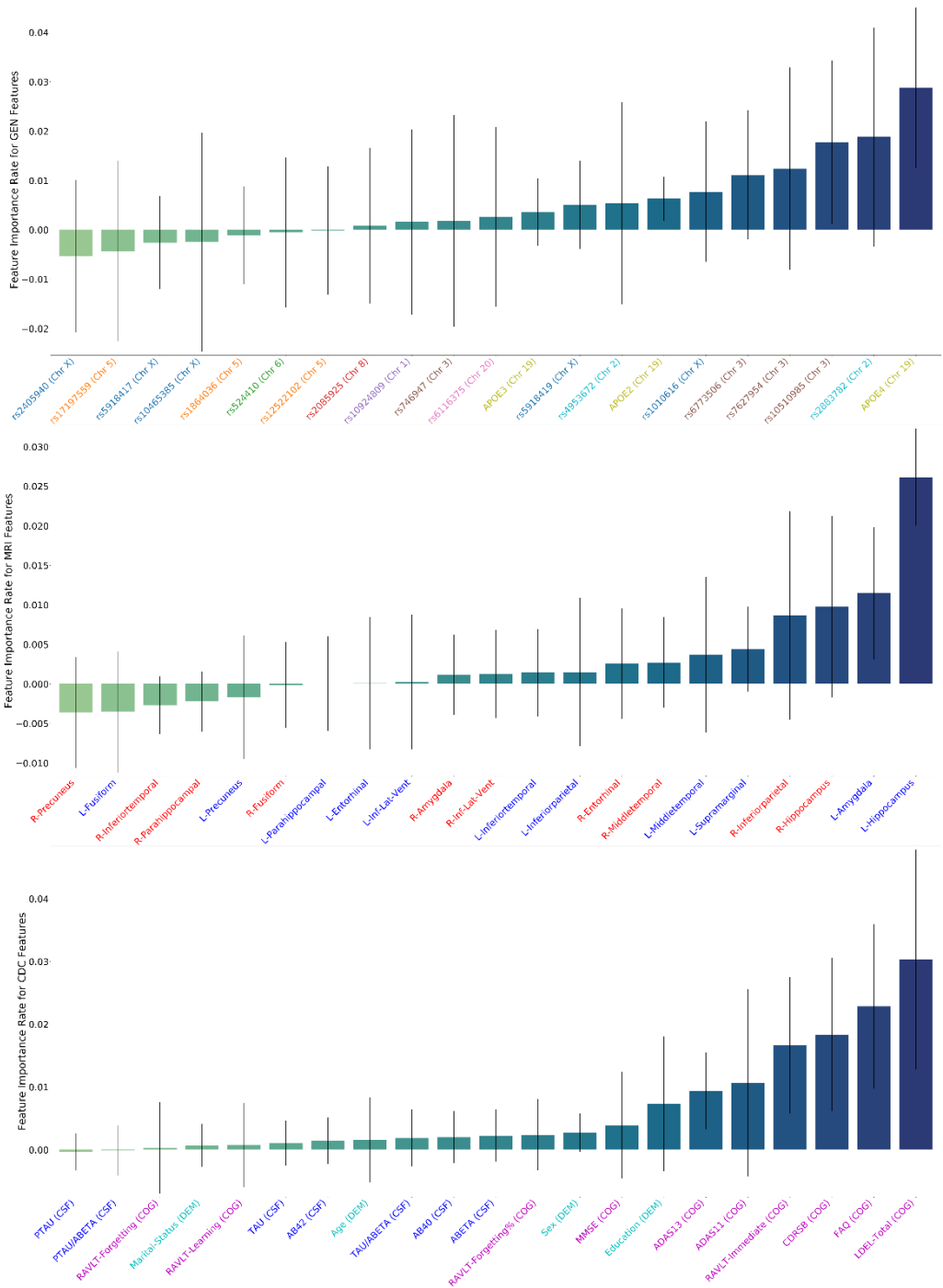

Figure 4. Feature importance results for GEN (top), MRI (middle), and CDC (bottom) feature sets. Each bar represents the mean score over ten splits, with the standard deviation indicated by a vertical error bar. X axis labels are color coded for ease of view. Colors on the top figure (GEN) indicate different chromosomes. On the middle figure (MRI), red shows right (R) and blue shows left (L) hemisphere of the brain. CSF, cognitive test (COG), and demographic (DEM) data are color coded on the bottom figure.

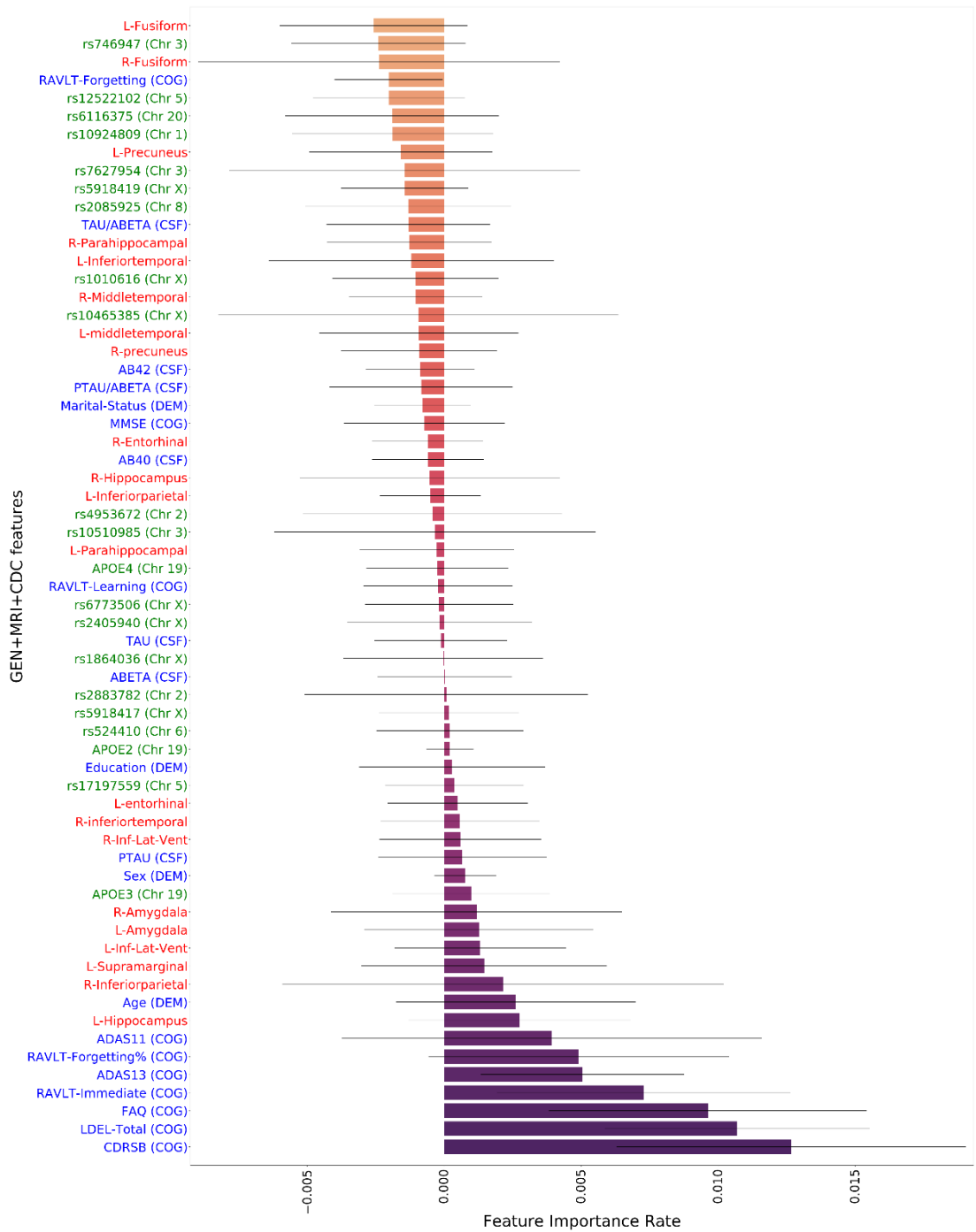

**Figure 5.** Feature importance results for the GEN+MRI+CDC feature set. Each bar represents the mean score over ten splits, with the standard deviation shown as a horizontal error bar. Data modalities have been color coded on y axis for ease of view (GEN, MRI, CDC). GEN labels indicate SNP (Chromosome number), MRI labels indicate Hemisphere-ROI name, and CDC labels indicate feature (data modality). DEM: demographic data; COG: cognitive test data.

**Table 5.** Performance comparison with $C^{td}$-index as evaluation method before and after removing features with negative importance using the Permutation Importance method. (the larger the better)

| Feature set (# step 1 / # step 2)[a] | $C^{td}$-index using all features (step 1 results) | $C^{td}$-index using features with $i_j > 0$ (step 2 results) |
|---|---|---|
| GEN (21/14) | 0.589 ± 0.049* | **0.628 ± 0.028** [b] |
| MRI (21/15) | 0.727 ± 0.025 | **0.744 ± 0.022** |
| CDC (21/19) | 0.822 ± 0.022 | **0.826 ± 0.027** |
| GEN+MRI+CDC (63/27) | 0.798 ± 0.026* | **0.831 ± 0.017** |

[a] (number of features in the main feature set / number of features with positive feature importance score),

[b] mean ± standard deviation over 10 splits, best performance in each row has been highlighted in **bold**.

* paired t-test with p < 0.05

**Table 6.** Performance comparison with IBS as evaluation method before and after removing features with negative importance using the Permutation Importance method. (the smaller the better)

| Feature set (# step 1 / # step 2)[a] | IBS using all features (step 1 results) | IBS using features with $i_j > 0$ (step 2 results) |
|---|---|---|
| GEN (21/14) | 0.206 ± 0.026 | **0.194 ± 0.020** [b] |
| MRI (21/15) | 0.160 ± 0.013 | **0.157 ± 0.016** |
| CDC (21/19) | **0.111 ± 0.019** | 0.116 ± 0.021 |
| GEN+MRI+CDC (63/27) | 0.122 ± 0.016 | **0.109 ± 0.014** |

[a] (number of features in the main feature set / number of features with positive feature importance score),

[b] mean ± standard deviation over 10 splits,

best performance in each row has been highlighted in **bold**.

Table 5 shows the impact of removing features with a negative feature importance rate from each feature set on the model's performance using $C^{td}$-index as the evaluation metric. As can be seen, removing features with negative importance rate has improved the performance in all cases, and this improvement was statistically significant ($p < 0.05$) for GEN and GEN+MRI+CDC feature sets. There were only two features in the CDC feature set with a negative importance rate, and removing them resulted in a slight improvement in performance. Table 6 displays a similar information, but this time using IBS as the evaluation metric.

## 4.3 Progressive vs. non-progressive survival analysis

Figure 6 displays the final survival probability across 10 splits for subjects in the progressive and non-progressive groups at the end of various time durations. A high survival probability indicates a high likelihood of NOT developing DAT or a low likelihood of developing DAT, whereas a low survival probability indicates a high likelihood of developing DAT after a certain time period.

One year after the initial visit (top left graph), approximately 74% of the non-progressive subjects had above 90% chance of survival while only 23% of the progressive subjects had this chance. Over a two-year period (top left graph), more than 85% of non-progressive subjects had a chance of survival greater than 50% (top 5 bars combined), where more than 58% of them had over 90% chance of survival. However, only half of the progressive subjects showed a survival chance of more that 50% with only 13% having over 90% survival probability.

The survival probability 5 years after the initial clinical visit is shown in the bottom left graph. As can be seen, approximately 40% of non-progressive subjects still had a 90% chance of survival, while less than 10% of progressive subjects had this chance. The proportion of progressive subjects with the lowest survival chance ($< 10\%$) was more than three times higher (34%) than the proportion of non-progressive subjects (10%) with the same survival probability. Ten years after the initial visit (bottom right graph), more than half of the non-progressive subjects still showed greater than 50% chance (bottom 5 bars combined) of survival, while about 27% showed a very low chance ($< 10\%$) of survival. In the progressive group, 78% of the subjects had a very low chance of survival ($< 10\%$), and only 8% had a chance of survival greater than 50%.

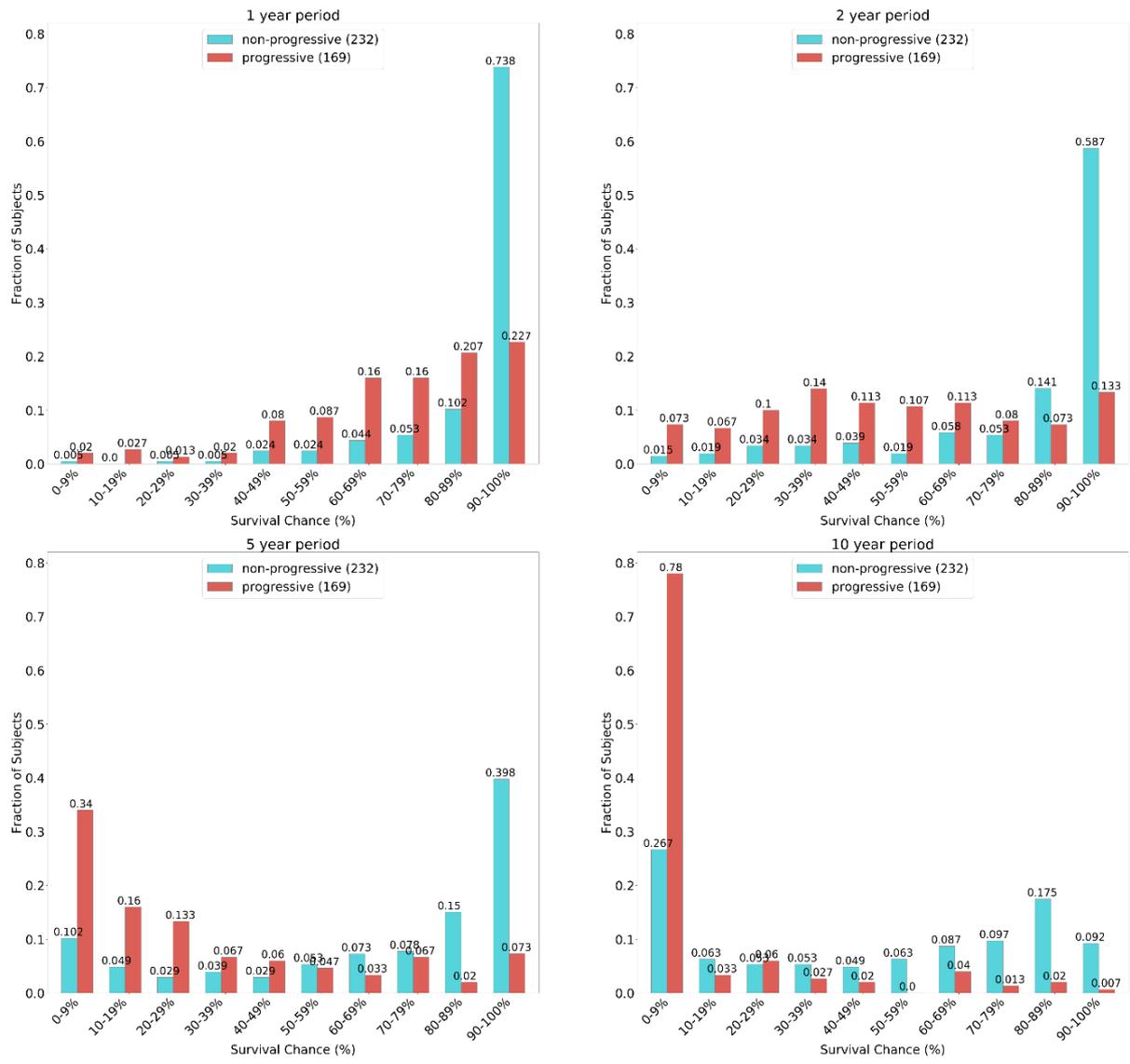

**Figure 6.  Estimated survival probability for progressive vs. non-progressive subjects over various time periods using the GEN+MRI+CDC feature set. The survival probability is divided into ten equal batches and each bar represents the proportion of subjects in each batch for both groups. The legend displays the total number of subjects in each group. The cyan bars represent data for non-progressive subjects, and the red bars show data for progressive subjects.**

The data related to the true time-to-conversion to Alzheimer's disease (event time) is available for progressive subjects, and can be compared to the predicted time-to-conversion.

Figure 7 shows a histogram of the differences between the predicted and true event times for progressive subjects (150 subjects) using the GEN+MRI+CDC feature set. Here, the predicted time is defined as the time when a subject's survival probability reaches 50%. If a subject's survival likelihood does not drop below 50% during the 10-year timeframe, it means the model regarded the subject to be at low risk of developing DAT and thus a non-converter based on the input information provided; otherwise, the subject is considered a converter. To obtain the time differences for all progressive subjects, we set the time-to-conversion as 20 years from the initial visit for those subjects whose survival probability did not reach 50% at the end of the 10-year period.

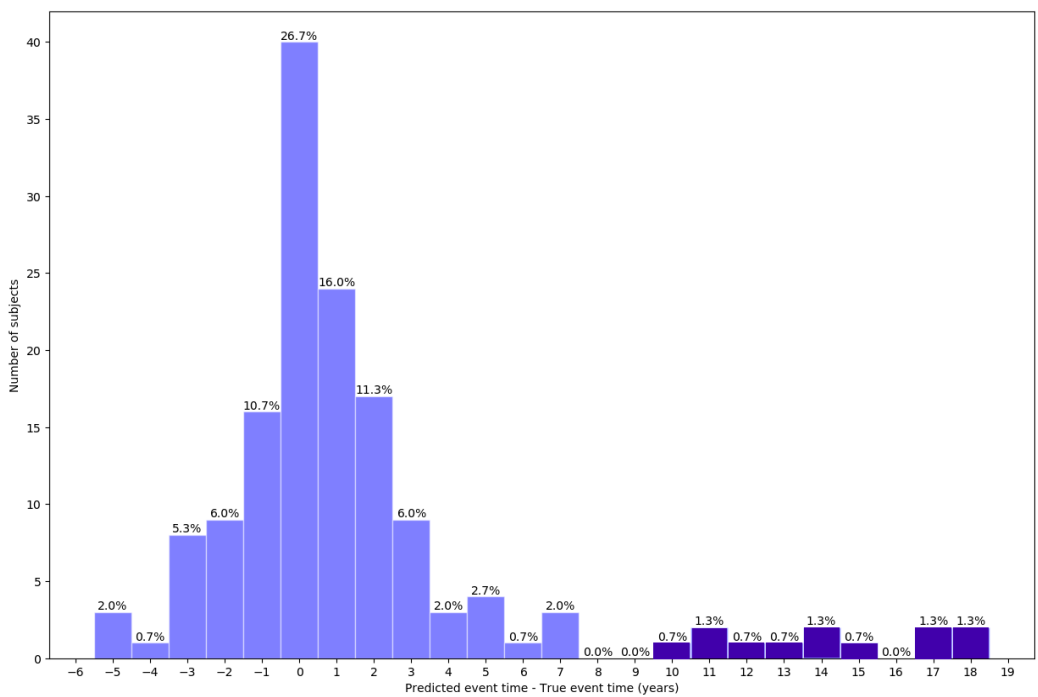

**Figure 7.** **Histogram of the difference between predicted and true event times for 150 progressive subjects using the GEN+MRI+CDC feature set. The predicted event time is the time when a subject's survival probability reaches 50%. If a subject's survival probability does not reach 50% by the end of the 10-year period, the subjects is considered a DAT non-converter (shown in dark purple). The percentage of subjects with a specific time difference is represented by a number printed above each bar.**

In Figure 7, each bar represents the number of subjects with a specific range of time difference. The bar at zero, for example, represents the number of subjects whose $predicted\ event\ time - true\ event\ time$ was between -0.5 and 0.5 years, and the bar at 2 shows the number of subjects with $predicted\ event\ time - true\ event\ time \in (1.5, 2.5)$. The true and predicted event times were very close to each other for 40 subjects (26.7%), with a difference of close to zero. In addition, more than half of the progressive subjects (80 out of 150, or 53.4%) had a time difference of less than 1.5 years, as shown by the three bars at -1, 0 and 1. The predicted event time was earlier than the actual event time for 37 subjects (24.7%; bars with negative time difference), indicating that our model detected the risk of developing DAT for those subjects prior to the actual event, which can be used as a biomarker in clinical settings. Despite the difference between the true and predicted event times, 92% of the subjects were correctly predicted as converters and only 8% (12 outlier subjects) of them were predicted as non-converters based on the survival data.

# 5 Discussion

## 5.1 Analyzing feature importance results

The order of feature importance rates shown in Figure 5 closely reflects the results of performance comparison using different feature sets (shown in Figure 3). Having 7 of the top 10 most important features from the CDC modality demonstrates the importance of these features on performance results and explains why adding CDC features to other feature sets results in an improved performance. Based on the results shown in Figure 5, we can rank the contribution of single modality feature sets as follows: GEN features contribute the least, followed by MRI features and CDC features, with CDC features contributing the most to the model performance. The same pattern can be seen in the performance comparison results between GEN (blue), MRI (orange), and CDC (green) feature sets displayed in Figure 3.

Cognitive test features (COG) were found to play an important role in our analysis. Our study included a total of 10 COG features; feature importance results using the CDC feature set declared all ten to be have a positive effect on the outcome (Figure 4, bottom graph), and feature importance results using all features declared seven to be important with an importance rate higher than all other features (Figure 5). CSF and genetic factors, on the other hand, were discovered to be less useful. We had 7 CSF and 21 GEN features in total. 14 of 21 GEN features (Figure 4, top graph) and 5 of 7 CSF features (Figure 4, bottom graph) were found to be important in single modality feature sets. The only two features with negative feature importance for CDC came from the CSF measures. Furthermore, the feature importance results using all features (Figure 5) only identified 6 GEN and 2 CSF features as significant. Our findings are important since it is much easier, cheaper, and less time consuming to perform cognitive tests than it is to acquire other data modalities, and our experiments demonstrate that cognitive tests are more promising, and can produce results that outperform those produced using other features.

## 5.2 Evaluation of time-to-conversion estimates and their benefits

Using survival analysis techniques to estimate time-to-conversion to Alzheimer's disease has the following advantages over other methods such as regression. First, there is no need to exclude subjects who did not develop DAT during the study period when using survival analysis which allows us to benefit from data collected from all study participants and ensure we do not accidentally bias the study by excluding these subjects. Second, survival analysis estimates the survival chance over time, giving us access to the survival probability at multiple time points and the overall trend of survival for each patient with respect to their input phenotype, genotype, and lifestyle information.

To evaluate the time-to-conversion estimates and to assess their applicability in a clinical setting, we randomly selected four subjects from each of the non-progressive and progressive groups and displayed their predicted survival times along with the true censoring or event times in Figure 8. Figures A-D (left side) belong to the non-progressive group, and their censoring time (green line) has been displayed along with the estimated survival probability, whereas figures E-

H (right side) belong to the progressive group, and their event time (red line) has been displayed along with the estimated survival probability.

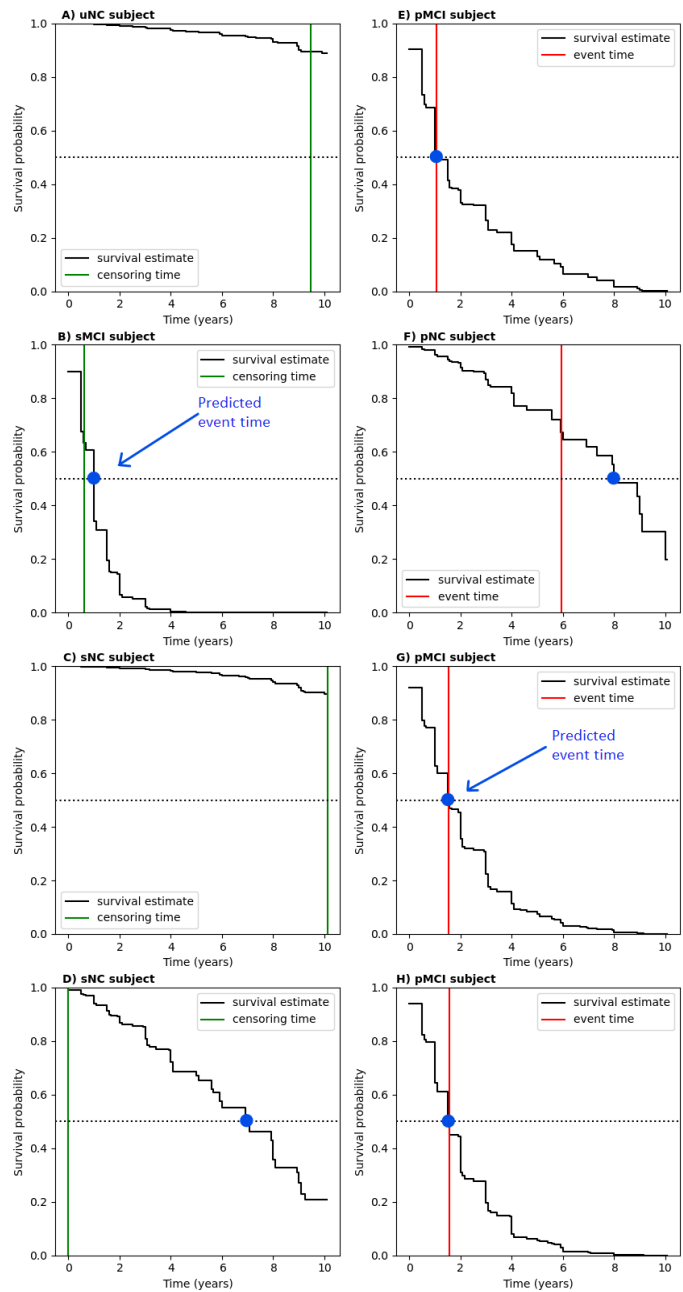

**Figure 8.** Comparison between the predicted survival estimates vs. actual censoring times (left) and event times (right) for 8 random subjects. The predicted event time is defined as the time when a subject's survival probability reaches 0.5. The horizontal dotted line represents a survival probability of 0.5, and the intersection of this line and the survival estimate curve (shown with a filled circle) represents the predicted event time for the subjects.

The uNC subject in A and the sNC subjects in C were censored around 10 years after their initial visit. At the end of the time window, both subjects still showed a very high survival probability, indicating they are at low risk of developing DAT. The sMCI subject in B was censored 6 months after the initial visit and had a survival probability of around 60% at the time of censorship. After 4 years, this subject's survival chance was dropped to 0%, indicating that the subject is likely to develop DAT in the future. The sNC subject in D was censored immediately after the initial visit and had a survival probability of 100% at the time of censoring. The subject's chance of survival decreased over time, reaching 20% at the end of the time window, indicating that they may develop DAT 7 to 10 years after their initial visit.

The pMCI subjects in graphs E, G, and H were all diagnosed with Alzheimer's disease around the same time, between one to two years after their initial visit. All three sMCI subjects had a similar survival curve, which could be translated into a similar disease severity. The pNC subject in F developed DAT symptoms 6 years after the initial visit. Despite having a higher chance of survival than other progressive subjects at event time, this subject's survival probability drops from 100% to 20% over a 10-year period, placing the subject in the high-risk category.

It is important to note that for all patients, only the information obtained at the initial clinical visit (time=0) was used as covariates in our model. For example, for pNC subjects, the model was only exposed to information when those subjects were completely healthy. This is clinically relevant because practitioners only have access to the information gathered at the current time point when at-risk patients visit them for the first time. Being able to accurately estimate the survival chance over time using only baseline information is extremely valuable because it provides practitioners with deep insight and enough time to plan appropriate care for each patient based on their future survival probability. Furthermore, while there is no cure for Alzheimer's disease at this time, many research facilities around the world are working around the clock to find a solution to permanently combat the disease. Accurate time-to-conversion estimates for at-risk patients can provide novel and potentially critical information for drug trials and the development of preventative measures. It can also aid in the selection of the appropriate cohort of patients for clinical trials, which can lead to a more promising outcome.

## 5.3 Evaluate the predicted time-to-conversion for pNC vs. pMCI subjects using different feature sets

The progressive group includes subjects from the pNC and pMCI stratified groups. The pNC subjects are those who were healthy at their initial clinical visit but developed DAT at a later point in time. As a result, all of the MRI, CSF, and cognitive test data gathered during the initial visit correlates with their healthy condition at the time. Genetic data, on the other hand, remains consistent over time and is thus the only factor that may point to these subjects' potential risk of developing Alzheimer's disease. The pMCI subjects are those who had MCI at their initial clinical visit and developed DAT at a future timepoint. In addition to genetic data, MRI, CSF, and cognitive test data may provide useful information about the disease stage in these subjects.

Figure 9 shows the difference between the predicted and true event times for subjects in the pNC (left; 13 subjects) and pMCI (right; 137 subjects) stratified groups using single modality (GEN, MRI, and CDC) feature sets as well as the combined data modality (GEN+MRI+CDC) feature set. As expected, for pNC subjects, using GEN features (A, left side) resulted in the most accurate prediction for time-to-conversion in comparison to using other feature sets. More than half of the pNC subjects (53.9%) had a time difference of less than 1.5 years. There were no outliers when GEN features were used, and all 13 subjects were correctly predicted to be DAT converters. Using MRI features (B, left side), 6 pNC subjects (46.2%) were correctly predicted as converters, while the remaining subjects were predicted as non-converters. Using CDC features (C, left side), only 1 pNC subject was correctly predicted as a converter, while using combined features (D, left side), 4 subjects were predicted as converters. Overall, GEN features were found to be the most helpful for in time-to-conversion prediction for pNC subjects, while CDC features were found to be the least helpful.

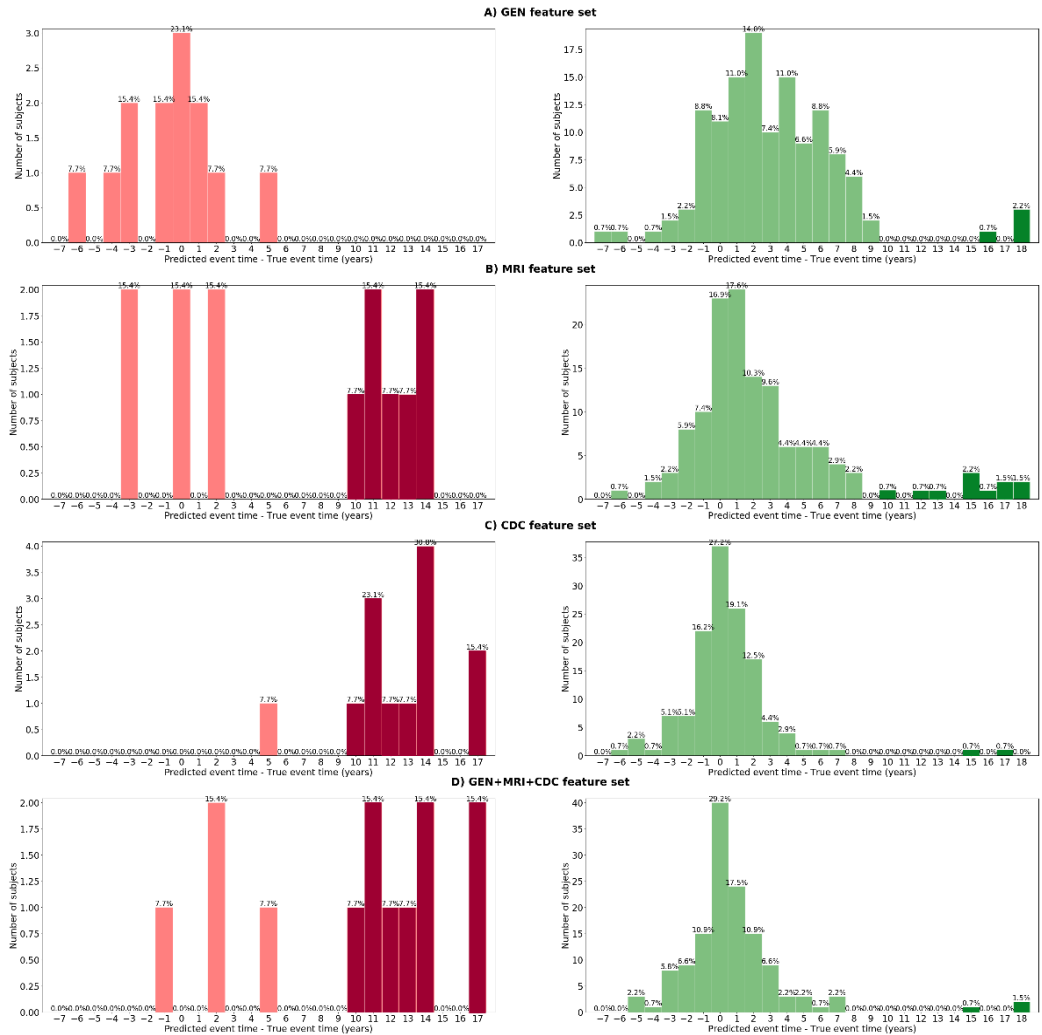

**Figure 9.** Histogram of the difference between the predicted and true event times for pNC (13 subjects, left) and pMCI (137 subjects, right) groups using A) GEN, B) MRI, C) CDC, and D) GEN+MRI+CDC feature sets. The predicted event time is the time a subject's survival probability reaches 50%. If a subject's survival probability does not reach 50% by the end of the 10-year period, the subjects is considered a DAT non-converter (shown in dark red and dark green for pNC and pMCI respectively). The percentage of subjects with a specific time difference is represented by a number printed above each bar.

For pMCI subjects, using CDC features (C, right side) resulted in the best time-to-conversion prediction among the single modality feature sets. 62.5% of pMCI subjects had a time difference of less than 1.5 years. There were only two outliers (1.4%), and 98.6% of the subjects were correctly classified as DAT converters. Using MRI features (B, right side) resulted in 11 outliers (8%), whereas using GEN features (A, right side) resulted in 4 outliers (2.9%). Although using GEN features resulted in fewer outliers, the time differences were more scattered when compared to using MRI features. Using combined features (D, right side) produced a histogram that was similar to the histogram produced by CDC features, indicating that CDC features are the most effective features in time-to-event prediction for pMCI subjects. Overall, all three data modalities were shown to be useful in time-to-event prediction for pMCI subjects, with CDC having the greatest contribution.

We discovered that genetic data (GEN) has the potential to detect the risk of developing DAT in the future for currently normal subjects, while other modalities have a lower predictive power. This information can be used in a clinical setting to determine the order of data acquisition for patients at different stages of the disease. Cognitive tests are cheap and fast to acquire and can be gathered first to determine the state of a patient's health. If a patient is determined to be healthy based on the results of cognitive tests at the initial clinical visit, genetic data can be obtained to determine the likelihood of developing DAT. Other data modalities, such as MRI and CSF data, can be collected during subsequent follow-up visits to closely study the patients' disease progression.

## 5.4 Limitations

Our study has some limitations: a) our results are limited by the sample size and characteristics of the group of subjects selected from the ADNI database. An approach to address this limitation would be to increase the sample size by using large-scale AD related databases such as UK Biobank database (https://www.ukbiobank.ac.uk), which can lead to a more robust model, b) the standard deviation appears to be high for all cases in Figures 4 and 5, which may be an indication of the Permutation Importance method's high sensitivity to the input data. The Permutation Importance approach provides a highly compressed, global insight into the model's behaviour and, to the best of our knowledge, is the best model inspection technique for back-box estimators. However, the method is highly dependent on both the main feature effect and the interaction effects with other features, and thus estimates how important a feature is for a specific model by taking into account the interaction between features. As a result, when translating feature importance using this method, it's is important to consider that changing the feature set used to train the model can affect the order of importance for features.

# 6 Acknowledgements

Funding for this research is gratefully acknowledged from Alzheimer Society Research Program, National Science Engineering Research Council (NSERC), Canadian Institutes of Health Research (CIHR), Fondation Brain Canada, Pacific Alzheimer's Research Foundation, the Michael Smith Foundation for Health Research (MSFHR), the National Institute on Aging (R01 AG055121-01A1, R01 AG069765-01, R01 AG071514-01), National Institute of Neurological Disorders and Stroke (NINDS) (R01 NS101483- 01A1), Precision Imaging Beacon, University of Nottingham, Canadian Statistical Sciences Institute (CANSSI), and Wake Forest School of Medicine Start-up Funds. We thank Compute Canada for the computational infrastructure provided for the data processing in this study.

Data collection and sharing for this project was funded by the Alzheimer's Disease Neuroimaging Initiative (ADNI) (National Institutes of Health Grant U01 AG024904) and DOD ADNI (Department of Defense award number W81XWH-12-2-0012). ADNI is funded by the National Institute on Aging, the National Institute of Biomedical Imaging and Bioengineering, and through generous contributions from the following: AbbVie, Alzheimer's Association; Alzheimer's Drug Discovery Foundation; Araclon Biotech; BioClinica, Inc.; Biogen; Bristol-Myers Squibb Company; CereSpir, Inc.; Cogstate; Eisai Inc.; Elan Pharmaceuticals, Inc.; Eli Lilly and Company; EuroImmun; F. Hoffmann-La Roche Ltd and its affiliated company Genentech, Inc.; Fujirebio; GE Healthcare; IXICO Ltd.; Janssen Alzheimer Immunotherapy Research & Development, LLC.; Johnson & Johnson Pharmaceutical Research & Development LLC.; Lumosity; Lundbeck; Merck & Co., Inc.; Meso Scale Diagnostics, LLC.; NeuroRx Research; Neurotrack Technologies; Novartis Pharmaceuticals Corporation; Pfizer Inc.; Piramal Imaging;

Servier; Takeda Pharmaceutical Company; and Transition Therapeutics. The Canadian Institutes of Health Research is providing funds to support ADNI clinical sites in Canada. Private sector contributions are facilitated by the Foundation for the National Institutes of Health (www.fnih.org). The grantee organization is the Northern California Institute for Research and Education, and the study is coordinated by the Alzheimer's Therapeutic Research Institute at the University of Southern California. ADNI data are disseminated by the Laboratory for Neuro Imaging at the University of Southern California.

# 7  Conflict of Interest

The authors have no conflict of interest to report.

# Supplementary Material: Predicting Time-to-conversion for Dementia of Alzheimer's Type using Multi-modal Deep Survival Analysis

## A  List of features included in the study

Table A.1 includes a list of all 63 features included in this study and their details. These features include volume w-score measurements from 21 ROIs (MRI features), 21 genetic features including 3 APOE alleles and 18 SNPs, 4 demographic features, 10 features from the cognitive tests, and 7 CSF features. Cognitive tests, Demographic, and CSF features (21 features) are called CDC in short.

**Table A.1.    List of 63 features included in the study**

| Feature name | Feature category | Data type | Number of missing data | Description |
| --- | --- | --- | --- | --- |
| Amygdala - Left[a] | MRI[b] | Continuous | 0 | Subcortical region in the left hemisphere |
| Amygdala - Right[a] | MRI | Continuous | 0 | Subcortical region in the right hemisphere |
| Entorhinal - Left | MRI | Continuous | 0 | Cortical region in the left hemisphere |
| Entorhinal - Right | MRI | Continuous | 0 | Cortical region in the right hemisphere |
| Fusiform - Left | MRI | Continuous | 0 | Cortical region in the left hemisphere |
| Fusiform - Right | MRI | Continuous | 0 | Cortical region in the right hemisphere |
| Hippocampus - Left | MRI | Continuous | 0 | Subcortical region in the left hemisphere |
| Hippocampus - Right | MRI | Continuous | 0 | Subcortical region in the right hemisphere |
| Inferior-parietal - Left | MRI | Continuous | 0 | Cortical region in the left hemisphere |
| Inferior-parietal - Right | MRI | Continuous | 0 | Cortical region in the right hemisphere |
| Inferior-temporal - Left | MRI | Continuous | 0 | Cortical region in the left hemisphere |
| Inferior-temporal - Right | MRI | Continuous | 0 | Cortical region in the right hemisphere |

| Feature name | Feature category | Data type | Number of missing data | Description |
| --- | --- | --- | --- | --- |
| Inferior-lateral-ventricle - Left | MRI | Continuous | 0 | Inferior or temporal horn of the lateral ventricle in the left hemisphere |
| Inferior-lateral-ventricle - Right | MRI | Continuous | 0 | Inferior or temporal horn of the lateral ventricle in the right hemisphere |
| Middle-temporal - Left | MRI | Continuous | 0 | Cortical region in the left hemisphere |
| Middle-temporal - Right | MRI | Continuous | 0 | Cortical region in the right hemisphere |
| Parahippocampal - Left | MRI | Continuous | 0 | Cortical region in the left hemisphere |
| Parahippocampal - Right | MRI | Continuous | 0 | Cortical region in the right hemisphere |
| Precuneus - Left | MRI | Continuous | 0 | Cortical region in the left hemisphere |
| Precuneus - Right | MRI | Continuous | 0 | Cortical region in the right hemisphere |
| Supramarginal - Left | MRI | Continuous | 0 | Cortical region in the left hemisphere |
| APOE-ε2 | GEN[b] | Binary | 0 | SNP on chromosome 19 Gene[c]: APOE |
| APOE-ε3 | GEN | Binary | 0 | SNP on chromosome 19 Gene: APOE |
| APOE-ε4 | GEN | Binary | 0 | SNP on chromosome 19 Gene: APOE |
| rs524410 | GEN | Categorical | 0 | SNP on chromosome 6 Gene: LOC112267968 |
| rs746947 | GEN | Categorical | 0 | SNP on chromosome 3 Gene: FRMD4B |
| rs1010616 | GEN | Categorical | 0 | SNP on chromosome X Gene: ZDHHC15 |
| rs1864036 | GEN | Categorical | 0 | SNP on chromosome 5 Gene: LOC105379004 |
| rs2085925 | GEN | Categorical | 0 | SNP on chromosome 8 Gene: TRAPPC9 |
| rs2405940 | GEN | Categorical | 0 | SNP on chromosome X Gene: SHROOM2 |

| Feature name | Feature category | Data type | Number of missing data | Description |
| --- | --- | --- | --- | --- |
| rs2883782 | GEN | Categorical | 0 | SNP on chromosome 2<br>Gene: MYO3B |
| rs4953672 | GEN | Categorical | 0 | SNP on chromosome 2<br>Gene: HAAO and MTA3 |
| rs5918417 | GEN | Categorical | 0 | SNP on chromosome X<br>Gene: SYTL5 |
| rs5918419 | GEN | Categorical | 0 | SNP on chromosome X<br>Gene: SYTL5 |
| rs6116375 | GEN | Categorical | 0 | SNP on chromosome 20<br>Gene: PRNP |
| rs6773506 | GEN | Categorical | 0 | SNP on chromosome 3<br>Gene: FRMD4B |
| rs7627954 | GEN | Categorical | 0 | SNP on chromosome 3<br>Gene: TNIK |
| rs10465385 | GEN | Categorical | 0 | SNP on chromosome X<br>Gene: LINC02154 |
| rs10510985 | GEN | Categorical | 0 | SNP on chromosome 3<br>Gene: FRMD4B |
| rs10924809 | GEN | Categorical | 0 | SNP on chromosome 1<br>Gene: CNST |
| rs12522102 | GEN | Categorical | 0 | SNP on chromosome 5<br>Gene: LOC105379004 |
| rs17197559 | GEN | Categorical | 0 | SNP on chromosome 5<br>Gene: LOC105379004 |
| Aβ40 (CSF[b]) | CDC[b] | Continuous | 41 | 40-residue Amyloid-β peptide |
| Aβ42 (CSF) | CDC | Continuous | 40 | 42-residue Amyloid-β peptide |
| Aβ (CSF) | CDC | Continuous | 176 | Amyloid-β |
| ptau (CSF) | CDC | Continuous | 176 | Phosphorylated Tau |
| ptau/Aβ (CSF) | CDC | Continuous | 176 | Phosphorylated Tau to Amyloid-β ratio |

| Feature name | Feature category | Data type | Number of missing data | Description |
|---|---|---|---|---|
| Tau (CSF) | CDC | Continuous | 176 | Tau protein |
| tau/Aβ (CSF) | CDC | Continuous | 176 | Total Tau to Amyloid-β ratio |
| Age (DEM[b]) | CDC | Continuous | 0 | Age at baseline |
| Sex (DEM) | CDC | Binary | 0 | Biological sex assigned at birth |
| Education (DEM) | CDC | Categorical | 0 | Education at baseline |
| Marital status (DEM) | CDC | Categorical | 0 | Marital status at baseline |
| ADAS11 (COG[b]) | CDC | Categorical | 0 | Alzheimer's Disease Assessment Scale – 11 tasks |
| ADAS13 (COG) | CDC | Categorical | 1 | Alzheimer's Disease Assessment Scale – 13 tasks |
| CDRSB (COG) | CDC | Categorical | 0 | Clinical Dementia Rating scale Sum of Boxes |
| FAQ (COG) | CDC | Categorical | 2 | Functional Activities Questionnaire |
| LDELTOTAL (COG) | CDC | Categorical | 0 | Logical Memory - Delayed Recall - Total Number of Story Units Recalled |
| MMSE (COG) | CDC | Categorical | 0 | Mini Mental State Exam |
| RAVLT-forgetting (COG) | CDC | Categorical | 1 | Rey Auditory Verbal Learning Test - Forgetting (trial 5 - delayed) |
| RAVLT-immediate (COG) | CDC | Categorical | 1 | Rey Auditory Verbal Learning Test - Immediate (sum of 5 trials) |
| RAVLT-learning (COG) | CDC | Categorical | 1 | Rey Auditory Verbal Learning Test - Learning (trial 5 - trial 1) |
| RAVLT-%forgetting (COG) | CDC | Categorical | 2 | Rey Auditory Verbal Learning Test - Percent Forgetting |

[a] Left: region on the left hemisphere of the brain, Right: region on the right hemisphere of the brain,

[b] CSF: feature from the Cerebrospinal fluid category, DEM: feature from the Demographic category, COG: feature from the cognitive test category, CDC: the combination of COG, DEM, and CSF data,

[c] For those SNPs that do not fall exactly on a particular gene, nearest genes have been reported.

# B  Method comparison for handling missing data

To evaluate our method of handling missing data, we ran additional tests using two different approaches and compared the results. Initially, we replaced missing values in each feature with an out of range value (Twala, Jones, and Hand 2008) equal to three times the maximum value of that feature. This replacement was performed prior to the pre-processing step. The following options were used to replace missing data:

Option 1:   Mean imputation: Replacing missing data for a feature with the feature's average value before the pre-processing step (Dziura et al. 2013),

Option 2:   Replacing a feature's missing data in two consecutive steps: a) before feature pre-processing with the feature's average value, and b) after pre-processing with an out of range value equal to three times the maximum value of that feature.

**Table B.1.**   Method comparison using $C^{td}$-index as evaluation metric for handling missing data.

| Feature set | Replacement with mean (Option 1) | Replacement with mean & $3 \times$ max (Option 2) | Replacement with $3 \times$ max (main) |
| --- | --- | --- | --- |
| GEN | 0.589 ± 0.049 | 0.589 ± 0.049 | 0.589 ± 0.049 |
| MRI | 0.727 ± 0.025 | 0.727 ± 0.025 | 0.727 ± 0.025 |
| CDC | 0.812 ± 0.022 | 0.803 ± 0.026[*] | **0.822 ± 0.022**[a] |
| GEN+MRI | 0.760 ± 0.045 | 0.760 ± 0.045 | 0.760 ± 0.045 |
| GEN+CDC | **0.822 ± 0.027** | 0.807 ± 0.025 | 0.812 ± 0.019 |
| MRI+CDC | 0.829 ± 0.031 | 0.815 ± 0.023[*] | **0.831 ± 0.028** |
| GEN+MRI+CDC | **0.824 ± 0.023**[*] | 0.788 ± 0.032 | 0.800 ± 0.027 |

[a]   mean ± standard deviation over 10 splits, best performance in each row has been highlighted in **bold**,

[*]   paired t-test with p < 0.05 when compared to the main method.

The results of method comparison for handling missing data using $C^{td}$-index is displayed in Table B.1. The main method performed the best for CDC and MRI+CDC feature sets, and the difference was significant when compared to option 2. Option 1 (replacing missing data with mean) performed best for the GEN+CDC and GEN+MRI+CDC feature sets, and the difference was significant for the GEN+MRI+CDC feature set. Overall, the main method and option 1 performed very similarly, with the main method slightly outperforming the latter. Option 2 had the worst overall performance in terms of $C^{td}$-index. Table B.2 displays a similar information, but this time using IBS as the evaluation metric.

**Table B.2** Method comparison using IBS as evaluation metric for handling missing data.

| Feature set | Replacement with mean (Option 1) | Replacement with mean & 3 × max (Option 2) | Replacement with 3 × max (main) |
|---|---|---|---|
| GEN | 0.206 ± 0.026 | 0.206 ± 0.026 | 0.206 ± 0.026 |
| MRI | 0.160 ± 0.013 | 0.160 ± 0.013 | 0.160 ± 0.013 |
| CDC | 0.119 ± 0.017 | 0.125 ± 0.017 | **0.111 ± 0.019**[a] |
| GEN+MRI | 0.148 ± 0.023 | 0.148 ± 0.023 | 0.148 ± 0.023 |
| GEN+CDC | **0.117 ± 0.018** | 0.128 ± 0.022 | 0.121 ± 0.016 |
| MRI+CDC | 0.108 ± 0.025 | 0.115 ± 0.019 | **0.105 ± 0.016** |
| GEN+MRI+CDC | **0.106 ± 0.011**[*] | 0.142 ± 0.015[*] | 0.122 ± 0.016 |

[a]   mean ± standard deviation over 10 splits, best performance in each row has been highlighted in **bold**,

[*]   paired t-test with $p < 0.05$ when compared to the main method.

# C  Pre-processing method comparison

To investigate the effects of different data pre-processing methods on model performance and to evaluate the method used in this study, we employed two additional pre-processing approaches and compared the results. The following approaches were used:

Option 1:   All binary, categorical, and continuous features were left unchanged, i.e. raw data was fed directly into the network,

Option 2:   Encoding categorical features with entity embedding (Guo and Berkhahn 2016) half the size of the number of categories, standardizing continuous features, and leaving binary features unchanged.

**Table C.1.    Pre-processing method comparison using $C^{td}$-index as evaluation metric.**

| Feature set | No processing (Option 1) | Standardization (cont), Entity embedding (cat) (Option 2) | Standardization (cont+cat) (main) |
|---|---|---|---|
| GEN | 0.568 ± 0.057 | **0.593 ± 0.036**[a] | 0.589 ± 0.049 |
| MRI | **0.729 ± 0.025** | 0.727 ± 0.025 | 0.727 ± 0.025 |
| CDC | 0.797 ± 0.030* | 0.786 ± 0.038* | **0.822 ± 0.022** |
| GEN+MRI | 0.755 ± 0.044 | 0.729 ± 0.031 | **0.760 ± 0.045** |
| GEN+CDC | 0.782 ± 0.064* | 0.783 ± 0.033* | **0.812 ± 0.019** |
| MRI+CDC | 0.830 ± 0.033 | 0.793 ± 0.028* | **0.831 ± 0.028** |
| GEN+MRI+CDC | **0.811 ± 0.019*** | 0.785 ± 0.029 | 0.800 ± 0.027 |

[a]    mean ± standard deviation over 10 splits, best performance in each row has been highlighted in **bold**,

*    paired t-test with p < 0.05 when compared to the main method.

**Table C.2.    Pre-processing method comparison using IBS as evaluation metric.**

| Feature set | No processing (Step 5-1) | Standardization (cont), Entity embedding (cat) (Step 5-2) | Standardization (cont+cat) (main) |
|---|---|---|---|
| GEN | 0.216 ± 0.025 | **0.196 ± 0.014**[a] | 0.206 ± 0.026 |
| MRI | **0.157 ± 0.011** | 0.160 ± 0.013 | 0.160 ± 0.013 |
| CDC | 0.123 ± 0.015 | 0.134 ± 0.011* | **0.111 ± 0.019** |
| GEN+MRI | 0.149 ± 0.026 | 0.150 ± 0.018 | **0.148 ± 0.023** |
| GEN+CDC | 0.138 ± 0.021 | 0.137 ± 0.019 | **0.121 ± 0.016** |
| MRI+CDC | 0.106 ± 0.014 | 0.120 ± 0.016* | **0.105 ± 0.016** |
| GEN+MRI+CDC | **0.118 ± 0.019** | 0.125 ± 0.014 | 0.122 ± 0.016 |

[a]    mean ± standard deviation over 10 splits, best performance in each row has been highlighted in **bold**,

*    paired t-test with p < 0.05 when compared to the main method.

Table C.1 compares pre-processing methods using the $C^{td}$-index. The main method outperformed option 1 and option 2 methods using CDC, GEN+MRI, GEN+CDC, and MRI+CDC feature sets. The difference was statistically significant in 2 out of 4 cases between the main and option 1 methods, and 3 out of 4 cases between the main and option 2 methods. Option 1

performed best with MRI and GEN+MRI+CDC feature sets, with the improvement in performance being small for MRI but statistically significant for GEN+MRI+CDC comparing to the main method. Option 2 performed best when using GEN features, but the difference was small when compared to the main method. The main method showed the overall best performance using $C^{td}$-index. According to the IBS results shown in Table C.2, the main method had the overall best performance. However, the values were closer to each other when compared to the $C^{td}$-index results.

## D  Performance comparison between different survival models

There are numerous continuous-time and discrete-time survival analysis models available in the literature, and several studies have attempted to compare their performance on real-life datasets (Spooner et al. 2020; Beaulac et al. 2020; Kvamme and Borgan 2019). Here, in order to evaluate the performance of our model, we have trained our data using the traditional Cox proportional hazard (CoxPH) model (Cox 1972) as benchmark, as well as several best-performing deep learning-based models in the literature and compared the results with the method discussed in this manuscript, DeepSurv. The following deep learning-based survival analysis models have been used to train the data with all 63 available features in this step:

1. DeepHit (Lee et al. 2018): DeepHit is a discrete-time survival analysis model that estimates the survival distribution's probability mass function (PMF) and combines the log-likelihood of the censored data with a ranking loss for improved discriminative performance.
2. Nnet-survival or Logistic-Hazard (Gensheimer and Narasimhan 2019): Logistic-Hazard is a discrete-time survival model which parametrizes the discrete-time hazard rate with a neural network to optimize the survival likelihood.
3. PC-Hazard (Kvamme and Borgan 2019): Piecewise Constant Hazard is a continuous-time survival model in which the hazard rate is assumed to be piecewise constant in predefined intervals. .

4. Cox-Time (Kvamme, Borgan, and Scheel 2019): Cox-time is a non-linear and non-proportional extension of the classic Cox regression model. The hazard ratio in the Cox-Time model is defined to include time as a regular covariate ($g(t,x)$), allowing for model interactions between time and other covariates.

Model comparison results using the $C^{td}$-index are shown in Table D.1. The traditional CoxPH model performed very similarly to our main model (DeepSurv; (Katzman et al. 2018)). Using the GEN and GEN+MRI+CDC feature sets, CoxPH outperformed DeepSurv statistically. DeepSurv outperformed CoxPH in 4 of 7 cases, but the differences were not statistically significant. When compared to other deep learning-based models, DeepSurv performed the best across all feature combinations. All of the differences were statistically significant when compared to the Logistic-Hazard model, and almost all of the differences were statistically significant when compared to the PC-Hazard model, with the exception of when the GEN+MRI+CDC feature set was used. DeepHit and Cox-Time model outcomes were more comparable to DeepDurv outcomes. DeepSurv outperformed the DeepHit model statistically in 4 of 7 cases, while the results were significantly better in 2 of 7 cases when compared to Cox-Time.

Table D.2 displays the model comparison results using IBS as evaluation metric. The traditional CoxPH model performed very similarly to DeepSurv. DeepSurv outperformed CoxPH in 5 of the 7 cases, while CoxPH performed better in the remaining two. However, none of the differences were statistically significant. Our main model, DeepSurv, outperformed all deep learning-based models. However, contrary to the trend seen in Table 11, Logistic-Hazard and PC-Hazard performed most similarly to DeepSurv. DeepHit performed the worst in terms of IBS, and when compared to DeepSurv results, the differences were statistically significant in most cases, except when the GEN feature sets were used. When the MRI+CDC feature set was used, the IBS increase for the Cox-Time model was statistically significant compared to DeepSurv, and overall, Cox-Time was the second worst performing model.

**Table D.1.** Model performance comparison using $C^{td}$-index as evaluation metric. Here, performance of the main model, DeepSurv, is compared with the Cox proportional hazard model as benchmark, as well as four high-performing deep learning-based survival analysis models.

| Feature set | CoxPH (Cox 1972) | DeepHit (Lee et al. 2018) | LogisticHazard (Gensheimer and Narasimhan 2019) | PCHazard (Kvamme and Borgan 2019) | CoxTime (Kvamme, Borgan, and Scheel 2019) | DeepSurv (main (Katzman et al. 2018)) |
|---|---|---|---|---|---|---|
| GEN | **0.651** *(0.027)\** | 0.541 *(0.050)\** | 0.556 *(0.062)\** | 0.532 *(0.049)\** | 0.589 *(0.053)* | 0.589 *(0.049)* |
| MRI | **0.739** *(0.039)* | 0.702 *(0.044)\** | 0.669 *(0.061)\** | 0.686 *(0.037)\** | 0.718 *(0.040)* | 0.727 *(0.025)* |
| CDC | 0.817 *(0.028)* | 0.810 *(0.020)* | 0.763 *(0.049)\** | 0.779 *(0.027)\** | 0.811 *(0.036)* | **0.822** *(0.022)[a]* |
| GEN+MRI | 0.737 *(0.027)* | 0.736 *(0.050)* | 0.730 *(0.037)\** | 0.710 *(0.048)\** | 0.726 *(0.035)\** | **0.760** *(0.045)* |
| GEN+CDC | 0.811 *(0.022)* | 0.790 *(0.036)\** | 0.757 *(0.044)\** | 0.783 *(0.049)\** | 0.795 *(0.021)* | **0.812** *(0.019)* |
| MRI+CDC | 0.829 *(0.022)* | 0.814 *(0.023)\** | 0.775 *(0.051)\** | 0.770 *(0.051)\** | 0.799 *(0.026)\** | **0.831** *(0.028)* |
| GEN+MRI+CDC | **0.822** *(0.021)\** | 0.785 *(0.027)* | 0.773 *(0.040)\** | 0.784 *(0.040)* | 0.799 *(0.030)* | 0.800 *(0.027)* |

[a] mean *(standard deviation)* over 10 splits, best performance in each row has been highlighted in **bold**,

\* paired t-test with p < 0.05 when compared to the main method.

**Table D.2.** Model performance comparison using IBS as evaluation metric. Here, performance of the main model, DeepSurv, is compared with the Cox proportional hazard model as benchmark, as well as four high-performing deep learning-based survival analysis models.

| Feature set | CoxPH (Cox 1972) | DeepHit (Lee et al. 2018) | LogisticHazard (Gensheimer and Narasimhan 2019) | PCHazard (Kvamme and Borgan 2019) | CoxTime (Kvamme, Borgan, and Scheel 2019) | DeepSurv (main (Katzman et al. 2018)) |
|---|---|---|---|---|---|---|
| GEN | **0.197** *(0.009)*[a] | 0.216 *(0.011)* | 0.208 *(0.017)* | 0.208 *(0.012)* | 0.215 *(0.032)* | 0.206 *(0.026)* |
| MRI | 0.160 *(0.020)* | 0.186 *(0.010)** | 0.162 *(0.010)* | 0.162 *(0.009)* | 0.178 *(0.021)* | **0.160** *(0.013)* |
| CDC | 0.117 *(0.016)* | 0.150 *(0.016)** | 0.113 *(0.011)* | 0.113 *(0.012)* | 0.124 *(0.018)* | **0.111** *(0.019)* |
| GEN+MRI | 0.151 *(0.013)* | 0.180 *(0.014)** | 0.151 *(0.024)* | 0.150 *(0.016)* | 0.169 *(0.021)* | **0.148** *(0.023)* |
| GEN+CDC | 0.125 *(0.011)* | 0.148 *(0.015)** | 0.123 *(0.013)* | 0.122 *(0.013)* | 0.134 *(0.020)* | **0.121** *(0.016)* |
| MRI+CDC | 0.107 *(0.012)* | 0.145 *(0.013)** | 0.107 *(0.014)* | 0.107 *(0.010)* | 0.132 *(0.025)** | **0.106** *(0.016)* |
| GEN+MRI+CDC | **0.111** *(0.015)* | 0.157 *(0.016)** | 0.124 *(0.010)* | 0.123 *(0.015)* | 0.132 *(0.015)* | 0.122 *(0.016)* |

[a]    mean *(standard deviation)* over 10 splits, best performance in each row has been highlighted in **bold**,

\*    paired t-test with p < 0.05 when compared to the main method.